%% file: main.tex
\documentclass[10pt,twocolumn,letterpaper]{article}

\usepackage[applications]{wacv}      
\usepackage{wacv}              

\usepackage{epsfig}
\usepackage{graphicx}
\usepackage{amsmath}
\usepackage{amssymb}
\usepackage{booktabs}
\usepackage{appendix}
\usepackage{arydshln}
\usepackage{colortbl}

\usepackage{import}
\usepackage{xcolor}         
\usepackage{color,soul}     
\usepackage{makecell}
\usepackage{multirow}
\usepackage{enumitem}
\usepackage{graphicx}
\usepackage[accsupp]{axessibility}  

\definecolor{blue_munsell}{rgb}{0.36, 0.54, 0.66}
\definecolor{blue-violet}{rgb}{0.54, 0.17, 0.89}
\definecolor{byzantine}{rgb}{0.74, 0.2, 0.64}
\definecolor{caputmortuum}{rgb}{0.35, 0.15, 0.13}
\definecolor{other}{RGB}{83, 190, 198}
\definecolor{ashgrey}{rgb}{0.7, 0.75, 0.71}

\usepackage[pagebackref,breaklinks,colorlinks]{hyperref}

\newcommand\blfootnote[1]{%
  \begingroup
  \renewcommand\thefootnote{}\footnote{#1}%
  \addtocounter{footnote}{-1}%
  \endgroup
}

\newcommand*{\imagenet}{\textsc{ImageNet}\xspace}

\newcommand*{\deit}{\textsc{DeiT}\xspace}

\newcommand*{\deitsmall}{\textsc{DeiT-S}\xspace}

\newcommand*{\dino}{\textsc{Dino}\xspace}

\newcommand*{\jumpcp}{\textsc{JUMP-CP}\xspace}

\def\wacvPaperID{0545} 
\def\confName{WACV}
\def\confYear{2024}

\begin{document}

\title{Bridging Generalization Gaps in High Content Imaging Through Online Self-Supervised Domain Adaptation}

\author{Johan Fredin Haslum \textsuperscript{$1,2,3,$} 
\hspace{-1.5mm}
\thanks{Corresponding author: Johan Fredin Haslum \textless{}jhaslum@kth.se\textgreater{}} 
\hspace{0.05mm}
\thanks{These authors contributed equally to this work.  \vspace{-1.75mm}}
\qquad\hspace{-6mm}
Christos Matsoukas \textsuperscript{$1,2,3, \dagger$} 
\qquad\hspace{-6mm}
Karl-Johan Leuchowius \textsuperscript{$3$}
\qquad\hspace{-6mm}
Kevin Smith \textsuperscript{$1,2$}
\\\\
\textsuperscript{$1$} KTH Royal Institute of Technology, Stockholm, Sweden 
\textsuperscript{$2$} Science for Life Laboratory, \\ Stockholm, Sweden 
\textsuperscript{$3$}  AstraZeneca, Gothenburg, Sweden 
}

\maketitle

\blfootnote{ \textit{ \hrule \vspace{1mm} \hspace{-3.75mm}
Originally published at the Winter Conference on Applications of Computer Vision (WACV 2024).}}

\input{sections/0_Abstract}
\input{sections/1_Introduction}
\input{sections/2_Related_work}

\input{sections/3_Methods}
\input{sections/3.5_Experimental_Setup}
\input{sections/4_Experiments}

\input{sections/5_Discussion}
\input{sections/6_Conclusion}
\input{sections/7_acknowledgements}

{\small
\bibliographystyle{ieee_fullname}
\bibliography{References}
}

\newpage
\input{sections/9_Appendix}

\end{document}

%% file: sections/0_Abstract.tex
\begin{abstract}

High Content Imaging (HCI) plays a vital role in modern drug discovery and development pipelines, facilitating various stages from hit identification to candidate drug characterization. 
Applying machine learning models to these datasets can prove challenging as they typically consist of multiple batches, affected by experimental variation, especially if different imaging equipment have been used.
Moreover, as new data arrive, it is preferable that they are analyzed in an online fashion.
To overcome this, we propose CODA, an online self-supervised domain adaptation approach.
CODA divides the classifier's role into a generic feature extractor and a task-specific model. 
We adapt the feature extractor's weights to the new domain using cross-batch self-supervision while keeping the task-specific model unchanged.
Our results demonstrate that this strategy significantly reduces the generalization gap, achieving up to a 300\% improvement when applied to data from different labs utilizing different microscopes. 
CODA can be applied to new, unlabeled out-of-domain data sources of different sizes, from a single plate to multiple experimental batches.

\end{abstract}

%% file: sections/1_Introduction.tex
\section{Introduction}
\label{sec:intro}

High Content Imaging (HCI) plays a pivotal role in modern drug discovery and development, being used throughout preclinical drug discovery cascades. 
It can capture detailed phenotypic responses of cells treated with compounds or genetic perturbants, and reveal complex sub-cellular processes.
Machine learning can help analyze HCI data to unveil biological correlations, reveal mode-of-action, predict compound bioactivity, and predict toxicities \cite{schulze2013function, hofmarcher2019accurate, way2021predicting, NYFFELER2023116513, Haslum2023Bioact, simm2018repurposing}. 
Recent advances in ML for HCI have helped accelerate screening of compound libraries, enhanced data interpretation, and enabled novel therapeutic insights \cite{heiser2020identification}. 
However, several challenges hinder its full potential.

\begin{figure}[t]
\begin{center}
\includegraphics[width=1\columnwidth]{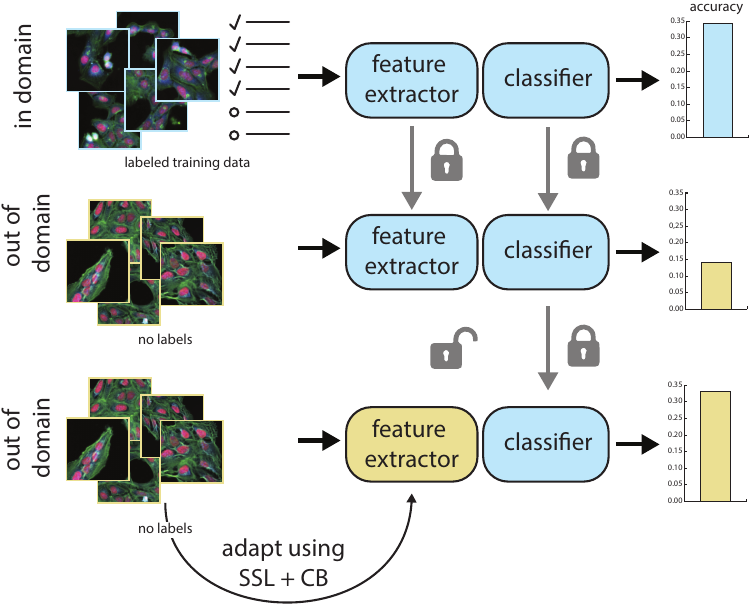}
\end{center}
\vspace{-2mm}
\caption{ \emph{Illustration of the proposed approach.  CODA divides the role of a classifier into two separate entities: one model designed to extract generic features and a subsequent task-specific model that operates on these features to accomplish the given task.
We then adapt the feature extractor when testing on a new source using self-supervision, while leaving
the task-specific model untouched. The result is a model that can be easily adapted to new out-of-domain data as it arrives, seeing a significant boost in performance, without the need for any labels.}
}
\label{fig:fig1}
\vspace{-4mm}
\end{figure}

In particular, the generalization gap presents significant challenges to HCI and drug discovery.
Discrepancies caused by variations in experimental conditions, apparatus, biological noise, and the presence of random or systematic errors can impede model performance. 
Limited ability of standard ML models to adapt or transfer across HCI settings results in reduced predictive accuracy \cite{Haslum2022MetadataguidedCL}. 
To fully harness the power of machine learning in HCI, there is a need for robust and adaptable models that can generalize effectively across different contexts and conditions without compromising performance. 

Recently, ten pharmaceutical companies, six supporting technology companies, and two non-profit partners formed the JUMP-CP (Joint Undertaking in Morphological Profiling) initiative to generate 
phenotypic response data for over 116,750 unique compounds, over-expression of 12,602 genes, and knockout of 7,975 genes using CRISPR-Cas9, all in human osteosarcoma cells (U2OS) \cite{JUMP_CP_DATASET}.
The dataset is estimated to be 115 TB in size and captures 1.6 billion cells and their single-cell profiles using the Cell Painting assay \cite{Bray2016CellPA,Cimini2022OptimizingTC}.
A subset of the compounds profiled by the consortium were profiled across all the participating labs, using the different equipment setups available in the individual labs. This dataset offers a unique opportunity to develop and test domain adaptation methods for HCI.

In this study, we leverage the newly-released JUMP-CP data to develop and validate a new approach for online self-supervised domain adaptation (SSDA), Cross-batch Online Domain Adaptation (CODA) which uses cross-batch self-supervision to adapt a feature extractor to incoming out-of-domain data.
By using the setup illustrated in Figure \ref{fig:fig1}, where the model is separated into an adaptable feature extractor and a frozen task-specific classifier, we are able realize huge improvements, up to 300\%, when the model is applied to data from different labs or different microscopes.
Crucially, this can be done without the need for any labels for the out-of-domain data.

We test our approach using data from different institutions in the JUMP-CP data repository -- training CODA using data from a source institution and performing SSDA to successfully adapt to the other institutions without access to any labels.
Our contributions can be summarized as follows:
\begin{itemize}
    \vspace{-1.5mm}
    \item Propose CODA, a self-supervised domain adaptation method, enabling online adaptation of a model trained on a single HCI data source to other out-of-domain sources (\eg, different institution or microscope) -- demonstrating its applicability to a variety of real-world experimental settings.

    \vspace{-1.5mm}
    \item Introduced ODA as an alternative approach when cross-batch consistency learning is not feasible, resulting in a slight performance drop from CODA but significant performance improvements over supervised methods.
    \vspace{-1.5mm}
    \item Conducted an extensive experimental validation on diverse subsets of data from the JUMP-CP repository, showcasing the robustness of the proposed approaches to variations in acquisition and apparatus, and verified the effectiveness of CODA in aligning the feature extractor to the target domain.
\end{itemize}

\noindent
The code to reproduce our experiments can be found at \href{https://github.com/cfredinh/coda}{https://github.com/cfredinh/coda}. 

%% file: sections/2_Related_work.tex
\section{Related Work}
\label{sec:related-work}

The problem of distributional shifts between the training and test sets is widely recognized to degrade performance in various domains \cite{Csurka2017DomainAF, Lee2022WildNetLD, Choi2021RobustNetID, Geirhos2018GeneralisationIH}. 
To mitigate these effects, traditional strategies involve gathering more data or employing sophisticated augmentation techniques to incorporate test distribution-like data into the training domain \cite{Hendrycks2018BenchmarkingNN, Zheng2016ImprovingTR}. 
However, these approaches may not always be feasible, as anticipating the expected domain shifts during testing is not always possible.
HCI data faces similar challenges due to domain shifts \cite{Chandrasekaran2020ImagebasedPF, Haslum2022MetadataguidedCL}.
Experimental batches in HCI data exhibit high homogeneity within themselves but have limited overlap with other batches due to inherent biological noise and variations in experimental setups. 
These variations are commonly referred to as \emph{batch effects} in HCI, representing undesirable domain shifts resulting from biological noise and difficult-to-control experimental conditions.

Research in the field of addressing distribution shifts focuses on two main directions: Domain Generalization (DG) and Domain Adaptation (DA). 
DG aims to learn domain-invariant features from one or multiple source domains during training, using techniques focused on identifying domain-invariant features \cite{Choi2021RobustNetID, Dou2019DomainGV, Pan2018TwoAO, Haslum2022MetadataguidedCL}. 
DA, on the other hand, leverages data from parts or the entire target domain during training, allowing for supervised or unsupervised alignment of features to handle distribution shifts \cite{Sun2016DeepCC, Peng2018MomentMF, yue2021prototypical, saito2020universal}. 
However, anticipating all possible distribution shifts during training is impractical, resulting in limited generalization capabilities across test domains. 
Consequently, performance cannot be guaranteed for unknown test domains.

The concept of updating model weights \emph{online} has recently gained attention, with \cite{sun2020test} introducing a pre-text task for weight updates, followed by \cite{Gandelsman2022TestTimeTW} using a image reconstruction task.
The underlying principle in most such approaches is that either only a sample or the full test set can be used to align the test domain without relying on and data associated with the primary task. 
This approach has shown clear performance gains in the natural image domain \eg~\cite{sun2020test, Gandelsman2022TestTimeTW}. 
Although these methods individually update weights for each sample, they are suboptimal for feature extraction in HCI data, as confirmed by our own experiments and recent findings \cite{SSL_JUMP}. 
Considering the nature of HCI data, which is often grouped into subsets such as wells, plates, or batches, adapting feature extraction strategies to these groups becomes an appealing and efficient option.

While test time domain adaptation has shown success in the Natural Imaging domain, its application in the medical and biomedical imaging domain, specifically in tasks like medical image segmentation \cite{Liu2021AdaptingOS} and image reconstruction \cite{Darestani2022TestTimeTC, He2021AutoencoderBS}, remains limited. 
Notably, there is a lack of research directly addressing domain shifts in medical and biomedical classification tasks, and particularly for HCI data, where these distributional shifts overwhelmingly dominate the learning signal.

%% file: sections/3_Methods.tex
\section{Methods}
\label{sec:methods}

In this study, we address the challenge of generalization gaps caused by domain shifts in new sources of High Content Imaging (HCI) data.
To tackle this problem, we propose a novel self-supervised domain adaptation (SSDA) strategy called CODA which is able to deal with the unique challenges associated with HCI data.

\begin{figure}[t]
\begin{center}
\includegraphics[width=1\columnwidth]{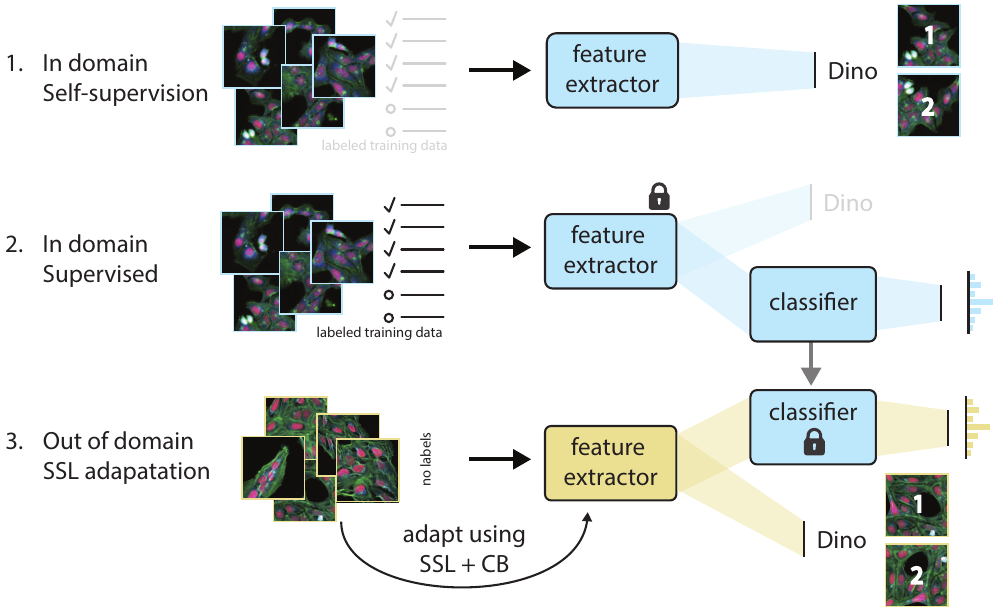}
\end{center}
\vspace{-2mm}
\caption{ \emph{The training strategy and deployment of the dual model in CODA. In the first step, a labeled data source is used to pretrain a feature extractor using self-supervision. In step 2, a classifier is appended to the feature extractor and supervised training is performed on the same data source. In step 3, the model is adapted online to a new unlabeled out-of-domain data source by using cross-batch consistency learning \cite{Haslum2022MetadataguidedCL} and self-supervision \cite{caron2021emerging} to adapt the feature extractor while keeping the weights of the classifier frozen.} }
\label{fig:fig2}
\vspace{-4mm}
\end{figure}

Building upon the recent work of   \cite{Gandelsman2022TestTimeTW}, we adopt a dual-model approach, which separates the model into a feature extractor and a classifier (Figure \ref{fig:fig1}).
Within our framework, the feature extractor is trained in a self-supervised manner and produces features which are then processed by the classifier to solve the classification task (Figure \ref{fig:fig2}).
Then, when unlabeled data from a new domain is encountered, one can update only the feature extractor using self-supervision to adapt to the new domain.
This  allows the model to adapt the feature extractor to the new domain while preserving the ability of the classifier to make correct predictions.

However, as demonstrated in this study, directly applying this design yields very poor performance in HCI data.
This is because the biological signals of interest are overshadowed by acquisition and experimental artifacts (see Table \ref{tab:main_results}). 
To overcome this obstacle, we made adaptations inspired by \cite{Haslum2022MetadataguidedCL} to modify the SSDA so that it becomes agnostic to these distracting artifacts.
This allows it to learn features that better distinguish the biological signal of interest and pass them on to the frozen classifier.

\vspace{-1mm}
\paragraph{Baseline}
The primary baselines we utilize in this study involve the supervised learning of a standalone Vision Transformer (ViT) model on HCI data, as commonly used. Once trained, we directly apply this model to the target dataset.

\vspace{-1mm}
\paragraph{The dual model}
In our approach, we utilize a vision transformer (ViT) model to learn meaningful representations from input patches. 
However, the standard ViT lacks the ability to differentiate generic low-level features from task-specific ones. 
To address this, we take inspiration from Test-Time Training (TTT) \cite{sun2020test, Gandelsman2022TestTimeTW}, which separates the model into a feature extractor and a classifier.
As seen in Figure \ref{fig:fig2}, the feature extractor learns generic representations through self-supervised training, while the classifier focuses on solving the specific task. 
In our study, we employ DINO \cite{caron2021emerging}, a consistency-based method, instead of reconstruction based Masked Autoencoders (MAE) \cite{Gandelsman2022TestTimeTW}. This is motivated by the subpar performance of MAE in HCI data \cite{SSL_JUMP} instead using DINO which has show better performance than other SSL approaches in HCI \cite{SSL_JUMP, Haslum2022MetadataguidedCL}. 
After the feature extractor is trained, it generates features that are used by the classification model.

\vspace{-1mm}
\paragraph{Self-supervised domain adaptation}
When a model is deployed in a new domain, the performance can be severely impacted by distribution shifts that alter the data representation, particularly in the case of HCI data, as discussed previously. 
These shifts primarily stem from intrinsic properties of the data rather than task-specific characteristics. 
In our problem the task remains constant, it is the appearance of the data that changes. 
Therefore, the domain adaptation method should focus on producing unbiased features that can be consumed by the classifier for the task at hand.

Adapting the features of a monolithic model to a new domain can be challenging due to the entanglement of low-level and high-level features, impacting both generic and task-specific representations. 
However, employing a dual model system allows for updating the feature extractor independently, while preserving the task-specific portion of the network. 
Inspired by this insight, we adopt the on-the-fly feature extractor update approach introduced by \cite{sun2020test, Gandelsman2022TestTimeTW}. 

Figure \ref{fig:fig2} illustrates the process. 
First, a labeled data source is utilized to pretrain the feature extractor through self-supervised learning (step 1). 
Subsequently, a classifier is appended to the feature extractor, and supervised training is conducted on the same data source (step 2). 
Finally, to adapt the model to a new unlabeled out-of-domain data source, using self-supervision techniques, specifically DINO \cite{caron2021emerging} are employed, allowing the feature extractor to be updated while keeping the classifier weights frozen (step 3).
By employing this approach, the classification model remains unaffected, while the feature extractor is adapted to extract generic representations suitable for the task, agnostic to the peculiarities of the data source.

\paragraph{Cross-batch consistency learning}
In High Content Imaging (HCI), the data collection process is characterized by discrete experimental batches, leading to distribution shifts caused by variations in experimental conditions, capturing settings, and time points. 
These shifts are commonly referred to as \emph{batch effects}. 
Ideally, each HCI image should capture only the biological effects of the treatment and no batch effects. 
However, in practice, batch effects often dominate the data, causing SSL methods to prioritize these confounding factors over the relevant biological signals. 
As a result, SSL methods tend to model batch effects rather than the desired biological signals, leading to suboptimal performance \cite{Haslum2022MetadataguidedCL}.

To address this challenge, Haslum et al. \cite{Haslum2022MetadataguidedCL} propose a solution called Cross-Domain Consistency Learning (CBCL), which builds upon the principles of consistency-based SSL methods. 
CDCL leverages the concept of consistent representation between pairs of images that share the same treatment but come from different domains. 
The underlying assumption is that when the network is presented with two images of the same treatment but from different batches, the shared signal of interest should be the biological signal rather than the batch effects. 
We adopt this strategy in the Self-Supervised in domain and adaptation step (steps 1 and 3 in Figure \ref{fig:fig2}) to mitigate the influence of batch effects and enhance the robustness of the self-supervised feature extractor.

\addtolength{\tabcolsep}{-5pt}  
\begin{table}[t]
\caption{
Imaging settings and data volume for the studied sources.
}
\vspace{-4mm}
\scriptsize
\input{tables/source_setting_table}
\label{tab:sample_setting}
\vspace{-4mm}
\end{table}
\addtolength{\tabcolsep}{+5pt}  

\begin{figure}[t]
\centering
\scriptsize
\begin{tabular}{@{}l@{\hspace{0.25mm}}c@{\hspace{0.25mm}}c@{\hspace{0.25mm}}c@{\hspace{0.25mm}}c@{\hspace{0.25mm}}c@{\hspace{0.25mm}}c@{\hspace{0.25mm}}l@{}}
\includegraphics[width=0.16\linewidth]{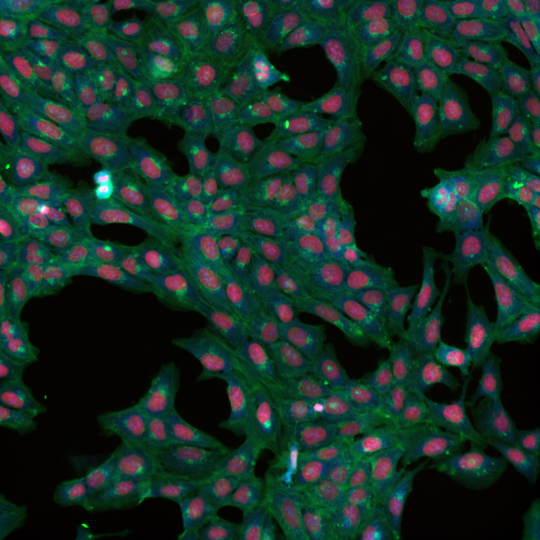} &
\includegraphics[width=0.16\linewidth]{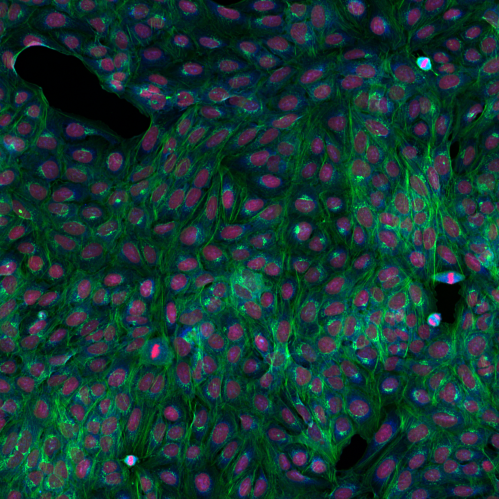} &
\includegraphics[width=0.16\linewidth]{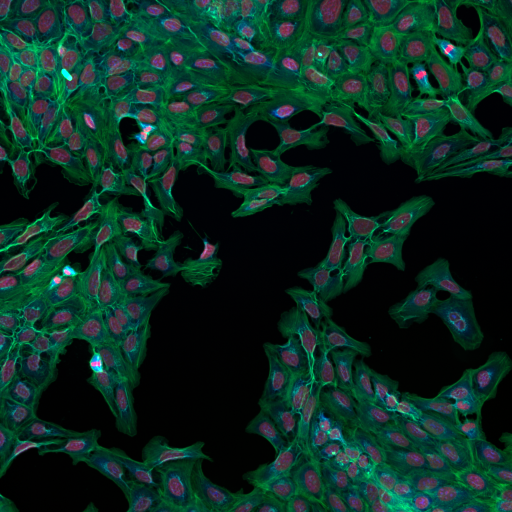} &
\includegraphics[width=0.16\linewidth]{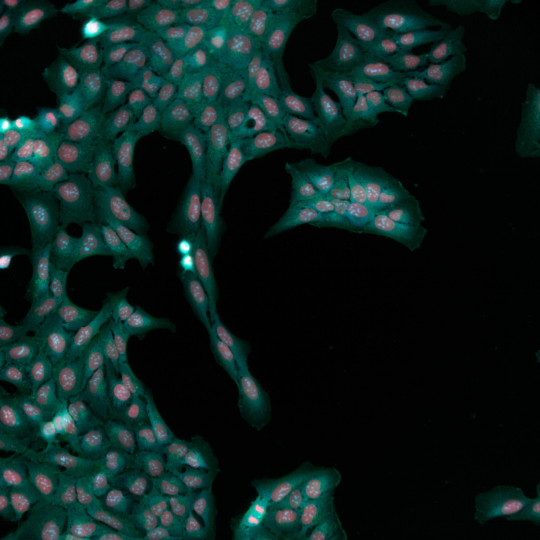} &
\includegraphics[width=0.16\linewidth]{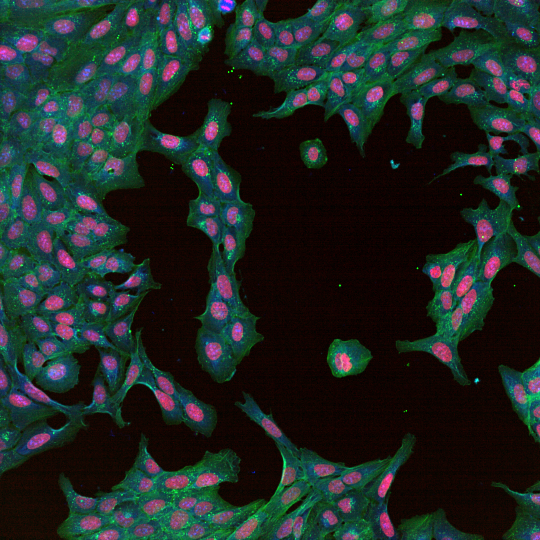} &
\includegraphics[width=0.16\linewidth]{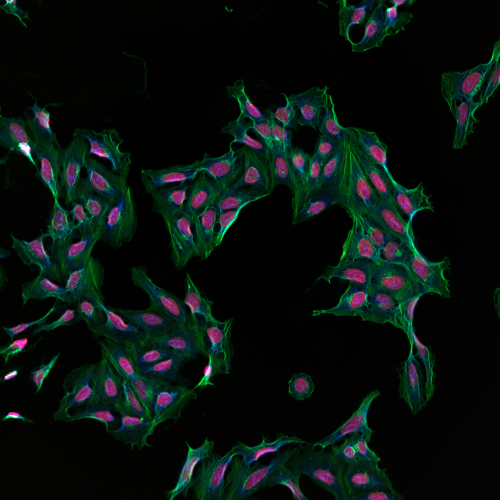} &
\\
\includegraphics[width=0.16\linewidth]{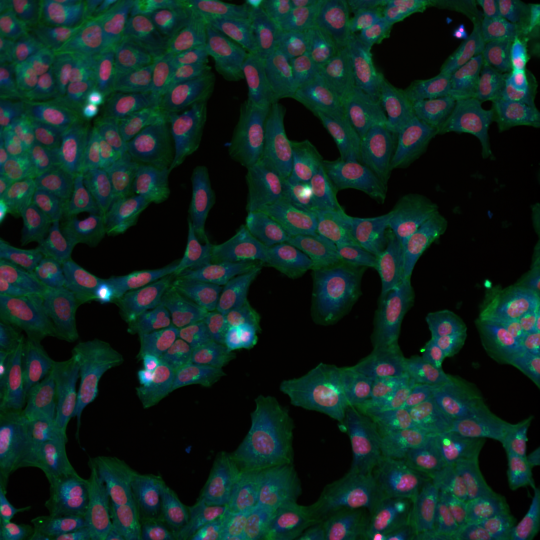} &
\includegraphics[width=0.16\linewidth]{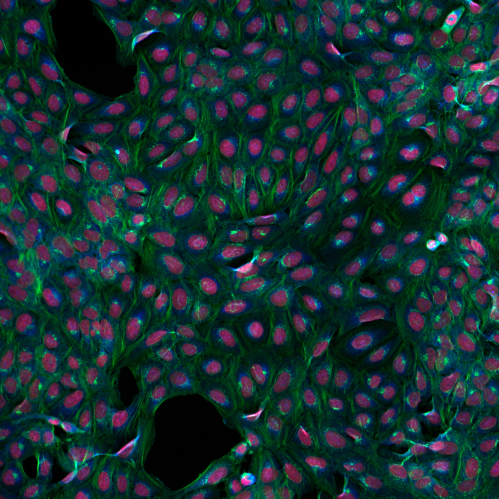} &
\includegraphics[width=0.16\linewidth]{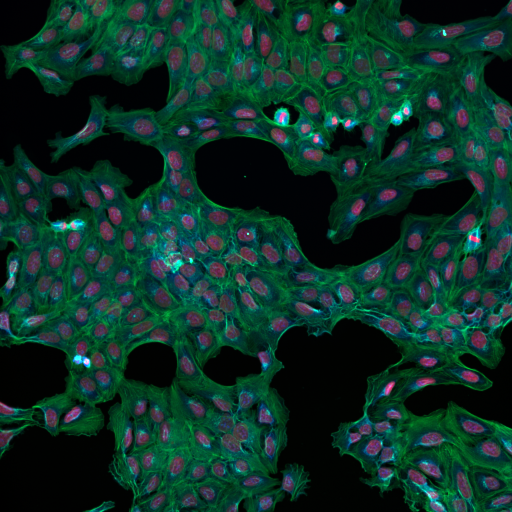} &
\includegraphics[width=0.16\linewidth]{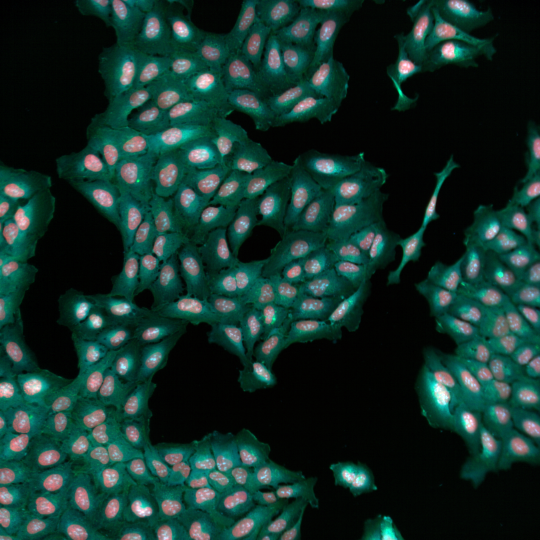} &
\includegraphics[width=0.16\linewidth]{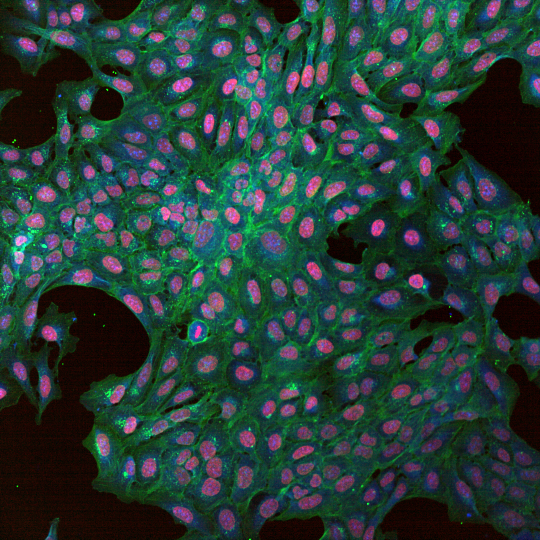} &
\includegraphics[width=0.16\linewidth]{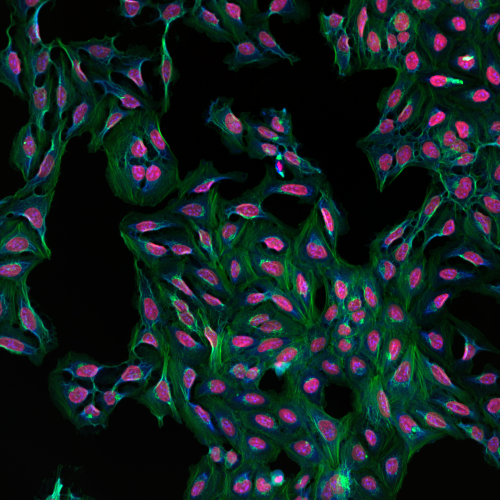} &
\\

\end{tabular}
\normalsize
\caption{
    \emph{Example images} from the six sources, showing the similarity between images of the same source but different compounds, highlighting the inherent variability and \textit{batch effects}. 
}
\label{fig:example_images_multi_compound}
\vspace{-4mm}
\end{figure}

\paragraph{Test Time Training}

Beyond the supervised baseline, we also include a recent Online Domain Adaptations approach, Test-Time Training (TTT) \cite{Gandelsman2022TestTimeTW}. TTT used a dual-model setup with a MAE \cite{he2022masked} as a feature extractor with a classification model stacked on top of it, these are trained in sequence. At test time the FE is update for each test image individually, see \cite{Gandelsman2022TestTimeTW} for more details.

%% file: tables/source_setting_table.tex
\begin{center}
\begin{tabular}{cccccc}
\toprule

\textbf{Source} & \textbf{Description} & \textbf{Objective} & \textbf{Batches} & \textbf{Plates} & \textbf{Images} \\
\midrule
S3 & Opera Phoenix, widefield, laser & 20X/1 & 13 & 25  & 85,409 \\
S5 & CV8000, confocal, laser & 20X/0.75 & 23 & 24 & 82,256 \\
S8 & ImageExpress Micro, confocal, LED & 20X/0.75 & 4 & 4 &  13,824 \\
S11 & Operetta, widefield, LED & 20X/1 & 4 & 7  & 23,373
\\
\cmidrule{1-6}
S7 & CV7000, confocal, laser & 20X/0.75 & 7 & 7  & 24,192 \\
S10 & CV8000, confocal, laser & 20X/0.75 & 6 & 6  & 13,812 \\

\bottomrule

\end{tabular}
\end{center}

%% file: sections/3.5_Experimental_Setup.tex
\section{Experimental Setup}
\label{sec:experimental_setup}

In this section, we describe the datasets, task used and the implementations details of the methods described in the previous section \ref{sec:methods}.
Starting with the task, we focus on Mechanism-of-Action (MoA) prediction.
The MoA of a compound describes how a substance produces a pharmacological effect, often involving determining which target, such as proteins or enzymes, the substance interacts with.
Understanding the MoA can help in predicting potential drug interactions, side effects, and can help guide the design of new, more effective drugs or therapies. 

\paragraph{Dataset} 

We conducted experiments using different subsets of the \jumpcp Cell Painting High Content Imaging set \cite{JUMP_CP_DATASET}. 
The \jumpcp dataset encompasses a wide range of compound and genetic perturbations that were imaged using an optimized version of the Cell Painting assay \cite{Bray2016CellPA, Cimini2022OptimizingTC}. 
This dataset was generated through collaborations between multiple institutions, making it an ideal choice for studying domain shifts due to its diverse origins and the inclusion of various microscope types and settings.
See 
Figure \ref{fig:example_images_multi_compound} 
for image examples.

For our analyses, we selected a subset of perturbations from all participating institutions of the \jumpcp initiative. 
This subset comprised 302 unique compounds that were imaged across 15 different centers, coming from the \jumpcp TARGET2 plates. 
It encompassed a total of 120 experimental batches and covered 141 distinct plates. 
As targets for our study, we aim to predict the Mechanism of Action (MoA) information associated with the compounds, which was obtained from the Drug Repurposing Hub \cite{Corsello2017TheDR}.
Among the compounds, 135 had single MoA labels, representing 54 unique MoA types. With the goal of predicting the MoA of each the compounds, we treat the problem as a multi-class classification task. 

Our main experiments focus on a subset of the data from four (anonymized) partners within the JUMP-CP consortium \cite{JUMP_CP_DATASET}. 
This subset consists of images captured using different microscopes and microscope types, with variations in objectives used. 
These four sources were selected as they were the largest subsets of data from each of the microscope types, providing the most diverse set of data sources, see top of Table \ref{tab:sample_setting} for details.
Two additional sources were also used for auxiliary testing, see Table \ref{tab:sample_setting}. 
These sources use similar microscope setups to the ones used in S5, allowing for comparison between similar image acquisition settings.
Additional details for the sources used in this work can be found in Appendix \ref{app:more_data}.

Note that in this work we use a subset of images with known MoA labels.
The raw image data were prepared using standard illumination correction and intensity outlier removal followed by downscale to half the original size and compressed, using \textit{DeepProfiler} \cite{DeepProfiler}. 
Finally, we reduce the image channels from five to the three most informative channels, based in observations by \cite{rxrx1experiment}.

\paragraph{Implementation details}

Throughout this study, we employ \deitsmall models \cite{Touvron2020TrainingDI}, initialized with pretrained weights from \imagenet \cite{Deng2009ImageNetAL}, similarly to \cite{matsoukas2023pretrained}. 
For all supervised models, we utilize cross-entropy loss for the MoA task. Our training process includes a linear warmup phase of 3 epochs, during which the learning rate gradually increases until it reaches a value of $10^{-4}$. 
Subsequently, we employ a step-wise learning rate reduction strategy, reducing the learning rate by a factor of 10 each time the validation loss and accuracy metric reach a saturation point.

To train the dual model, we follow a two-step approach depicted in Figure \ref{fig:fig2}. 
First, we perform a self-supervised step using \dino, adhering to the default settings described in \cite{caron2021emerging} with slight variations. 
This involves training for 300 epochs, using a learning rate of $10^{-4}$, linear warm-up for 10 epochs, followed by cosine annealing. 
We use an exponential moving average of 0.996, following the augmentation strategy from \cite{SSL_JUMP}. 
When incorporating cross-batch examples, the same setup is used, with the exception of how the augmented samples are combined, using one global and three local views from each images in the sampled pair. The pairs are sampled based on metadata, with the requirement that the images are of the same treatment, but come from distinct batches. 
In the second step, the supervised task-specific step, we stack a \deitsmall model on top of the frozen feature extractor trained in the first step. Passing the tokens of the feature extractor into the task-specific model by removing the embedding layer and replacing it with a linear layer.
The training strategy for this step remains consistent with the supervised baseline approach described earlier.

Finally, the third step is adaptation using DINO, with or without CB sampling, to adapt the feature extractor to new out-of-domain data. This is done following the exact same strategy as in the second step described above, unless otherwise stated, but applied to the test set images, crucially not relying on any labels related to the primary task. 

%% file: sections/4_Experiments.tex
\section{Experiments and Results}
\label{sec:experiments}

In previous sections, we discussed the significant impact of distribution shifts in HCI. 
In this section, we demonstrate how dramatically performance degrades when applying a standalone model to a new HCI source.  
We measured performance using Accuracy as the label proportions are maintained across sources.
We further report the F1 scores in Table \ref{tab:main_results_f1} in the Appendix.
We first assess the generalization and adaptation capabilities of standalone \deit models, which serve as our baselines. 
Next, we deploy the dual model and evaluate its performance \emph{with} and \emph{without} adapting the feature extractor on the test source. 
Finally, we incorporate online self-supervised domain adaptation to address batch effects and assess its effectiveness. 
The results of our main experiments can be found in Table \ref{tab:main_results} and Figure \ref{fig:fig3}.

\addtolength{\tabcolsep}{-5pt}  
\begin{table}[t]
\caption{Generalization performance across target sources (Acc.).
}
\vspace{-4mm}
\tiny
\input{tables/main_table_TTT}
\label{tab:main_results}
\vspace{-4mm}
\end{table}
\addtolength{\tabcolsep}{+5pt}  

In our experiments, we consider the following models and baselines: 
\begin{itemize}
\vspace{-1mm}
    \item \textbf{Supervised} A standalone \deitsmall trained in a supervised fashion on the source data, applied to domain-shifted target data.
\vspace{-1mm}    
    \item \textbf{Dual-model-DINO} A \deitsmall feature extractor with a \deitsmall classifier stacked on top. The feature extractor is self-supervised with DINO and the classifier is trained in a supervised manner, with the feature extractor being frozen, both on the source data.
\vspace{-1mm}    
    \item \textbf{Dual-model-CB} The same as above, but in addition to DINO we use cross-batch image pair sampling, as described in section \ref{sec:methods}. 
\vspace{-1mm}    
    \item  \textbf{ODA} Online domain adaptation -- this is the same as Dual-modal-DINO but the feature extractor is adapted to the target data using self-supervision.
\vspace{-1mm}    
    \item \textbf{TTT} Similar to ODA but using MAE instead of DINO 
    and updating the feature extractor one image at a time. 
\vspace{-5mm}    
    \item \textbf{CODA} Cross-batch Online Domain Adaptation -- this is the same as Dual-modal-CB but the feature extractor is adapted to the target data.
\end{itemize}

\noindent\textbf{Baseline performance} 
We begin our evaluation with the standalone \deitsmall classification models. 
When the models are trained and evaluated on the same data source, as shown in Table \ref{tab:main_results}, the performance ranges from 30.1 to 40.5 with a mean of 36.5\% when including all sources, in terms of MoA accuracy. 
This represents the situation when there is no domain shift and we have access to labels.
When these models are applied to other splits (introducing a domain shift and no access to labels), a significant drop in performance is observed, with values ranging from 5.9 to 19.5 (first row of each non-diagonal element of Table \ref{tab:main_results} and the first bar in Figure \ref{fig:fig3} in Appendix). 
The substantial domain shifts cause the models to be reduced in accuracy by up to 84\%.

\noindent\textbf{Performance of the dual model} 
Replacing the baseline with the dual model yields a performance similar to that of the standalone model on the in-domain data. 
Somewhat surprisingly, these models exhibit lower performance compared to the standalone models when applied to out-of-domain sources. 
The accuracy ranges from 4.7 to 14.8, as indicated in the second row of each non-diagonal element in Table \ref{tab:main_results} (and the second bar in Figure \ref{fig:fig3} in the Appendix).

\noindent\textbf{Performance when updating the feature extractor}
Both the standalone and dual models fail to generalize to new HCI data sources. 
However, the situation changes dramatically when we employ online domain adaptation (ODA) on the feature extractor of the dual model when we apply it to new sources. 
As illustrated Table \ref{tab:main_results} (fifth row of each non-diagonal element of the source split) and the fourth bar in  Figure \ref{fig:fig3} in the Appendix we observe a substantial performance improvement compared to both the standalone model and the dual model without updated feature extractor. 
The out-of-domain MoA accuracy for ODA ranges from 10.6 to 27.2 (a mean increase of $174.8\% \pm 32.5$ over the baseline) surpassing the performance of the standalone model -- although still not reaching the level achieved when testing and evaluating within the domain.

\noindent\textbf{Performance when employing cross-batch learning}
Introducing cross-batch consistency learning (CBCL) for self-supervision of the dual model's feature extractor brings additional improvements in both in-domain and out-of-domain scenarios. 
CODA combines CDCL with online domain adaptation (ODA), yielding an enormous performance boost over the baseline as 
shown in Figure \ref{fig:fig3} in the Appendix (last bar) and the last row of Table \ref{tab:main_results}.
CODA yields MoA accuracy ranging from 11.9 to 36.7, in many cases, the out-of-domain performance is comparable to the performance achieved when training and testing are conducted within the same domain -- in some cases even exceeding that (\eg S3$\rightarrow$S8 with CODA yields 33.9 while the S8$\rightarrow$S8 baseline is 31.0).
The average performance boost of CODA over the baseline is $232.6\% \pm 63.6$.
CDCL is also helpful when used without online domain adaptation, although it provides less of a performance boost than ODA. As seen in row three of the non-diagonals in Table \ref{tab:main_results} and the third bar in Figure \ref{fig:fig3} in the Appendix. 

\noindent\textbf{MAE Performance} 
The idea of using separate models for feature extraction and classification, followed by feature extraction alignment was inspired by \cite{Gandelsman2022TestTimeTW}, where they use MAE \cite{he2022masked} for self-supervision. 
However, the performance of MAE in HCI data have been shown to be inferior to DINO, while also it does not allow for training with cross-batch examples \cite{SSL_JUMP}.
For completeness, we report in Table \ref{tab:main_results}, Table \ref{tab:MAE_table} and in the Appendix the results when using MAE instead of \dino. 
In domain MAE performs slightly worse than \dino.
However, in the ODA and TTT setup MAE fails to approach \dino performance -- even without CB training.

\subsection{Analysis and Ablation Studies}
In this section, we delve into further analysis and conduct ablation studies to gain a deeper understanding of the performance of CODA and the other models introduced in our experiments.

\begin{figure}[t]
\begin{center}
\begin{tabular}{@{}c@{}}
\includegraphics[width=1\columnwidth]{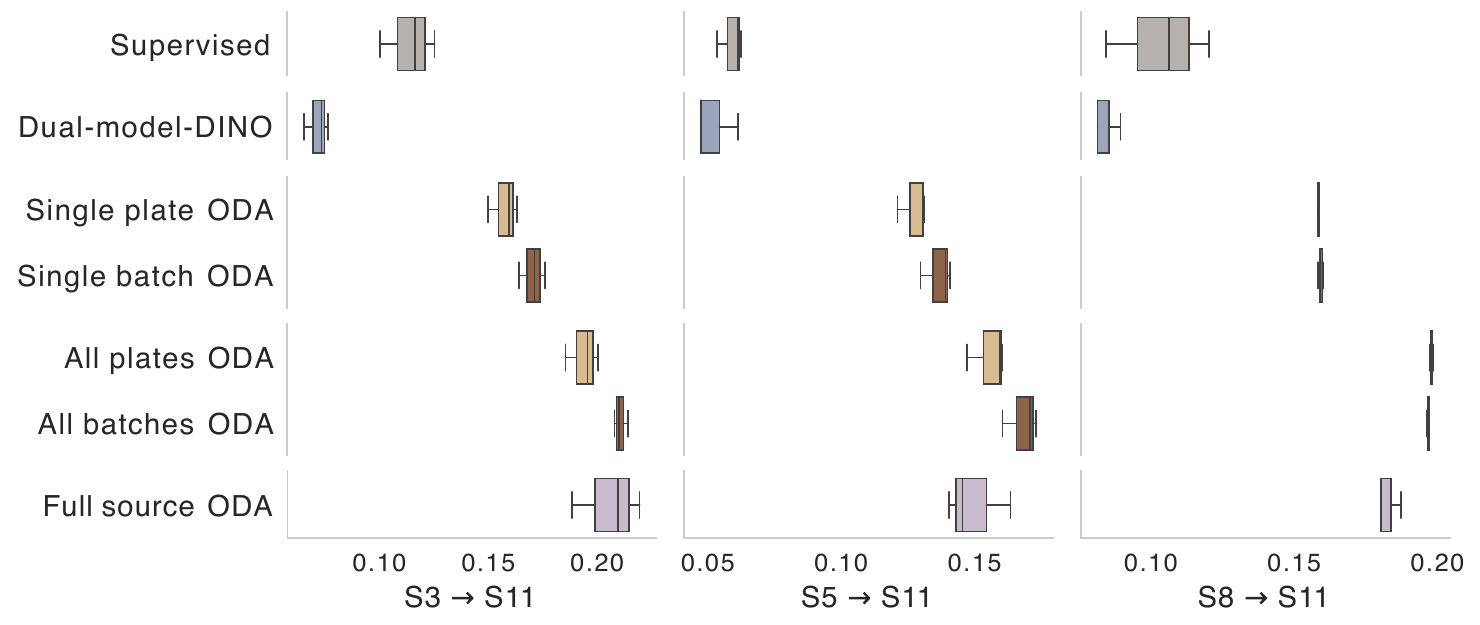}\\
\end{tabular}
\end{center}
\vspace{-4mm}
\caption{
    \emph{Effect of Data Granularity. (Accuracy)} 
    Investigating the performance impact when using ODA on different subsets (plate, batch, source) of the target domain. 
    \textit{Single plate/ batch} denotes the scenario where ODA is applied on a single plate or batch and then evaluated on the full target source.
    \textit{All plate/ batch} denotes the scenario where ODA is applied on a single plate or batch and tested on the same subset.
    }
\label{fig:abl}
\vspace{-4mm}
\end{figure}

\noindent\textbf{Generalization across similar microscopes} 
In order to gain further insights into the generalization capabilities of the different models, we conducted experiments using two additional data sources, S7 and S10, as test sets. 
These two sources share more similarities in terms of microscope setup with S5, compared with S3, S8, and S11. 
If the microscope setup was the primary factor influencing generalization, we would expect to observe improved performance as the similarity (domain distances) between the sources increases \cite{matsoukas2022makes}.
However, upon examining the results in Table \ref{tab:main_results}, we found no significant performance improvements for models trained on S5 and tested on S7 and S10 (S5$\rightarrow$S7 and S5$\rightarrow$S10). 
Surprisingly, the performance in S5$\rightarrow$S8 was actually higher, despite the use of distinct microscope types and illumination methods. 
This suggests that the impact of source-to-source variability on the model's generalization performance cannot be solely attributed to differences in imaging settings -- but rather to small differences in the protocol, reagents, or the environment.

\begin{figure}[t]
\begin{center}
\tiny
\arrayrulecolor{ashgrey}
\begin{tabular}{@{}c@{}|c@{}|c@{}}
Dual-Model-DINO (S3$\rightarrow$S3) & 
Dual-Model-DINO (S3$\rightarrow$S5) & 
ODA (S3$\rightarrow$S5) \\
\includegraphics[width=0.33\columnwidth]{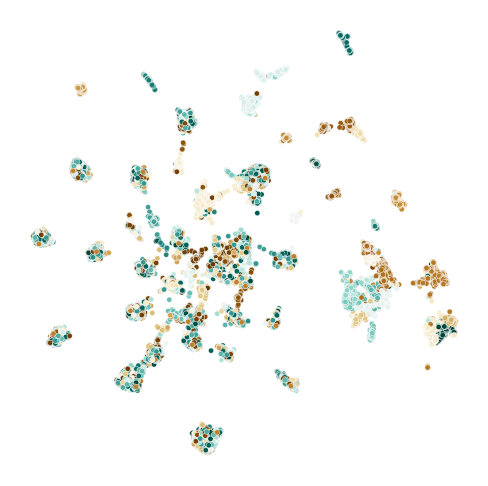}& 
\includegraphics[width=0.33\columnwidth]{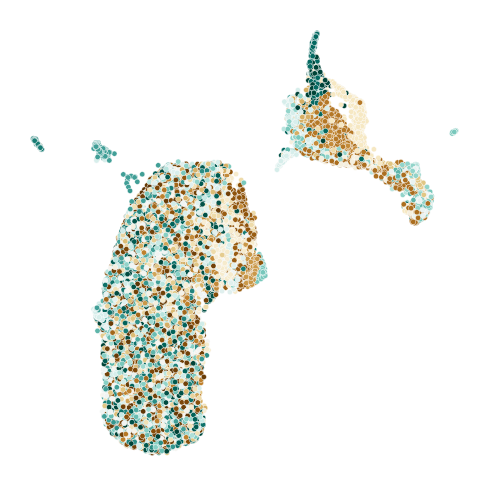}& 
\includegraphics[width=0.33\columnwidth]{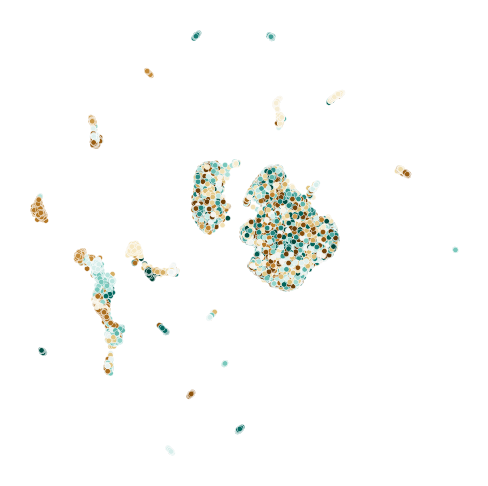}\\
\midrule
Dual-Model-CB (S3$\rightarrow$S3) & 
Dual-Model-CB (S3$\rightarrow$S5) & 
CODA (S3$\rightarrow$S5) \\
\includegraphics[width=0.33\columnwidth]{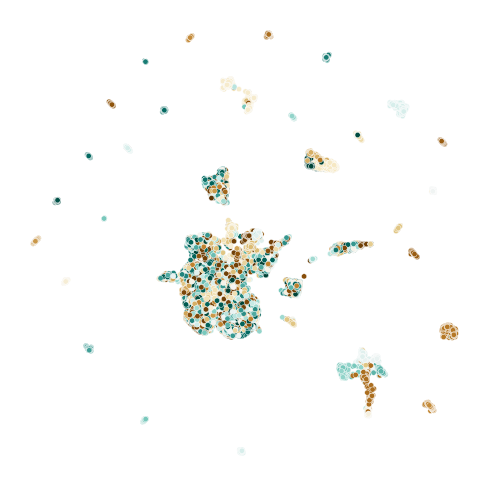}& 
\includegraphics[width=0.33\columnwidth]{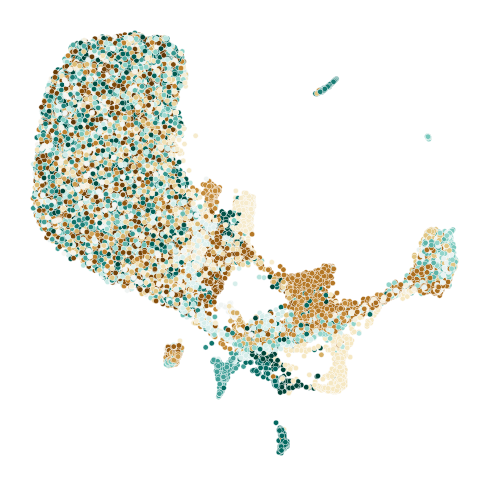}& 
\includegraphics[width=0.33\columnwidth]{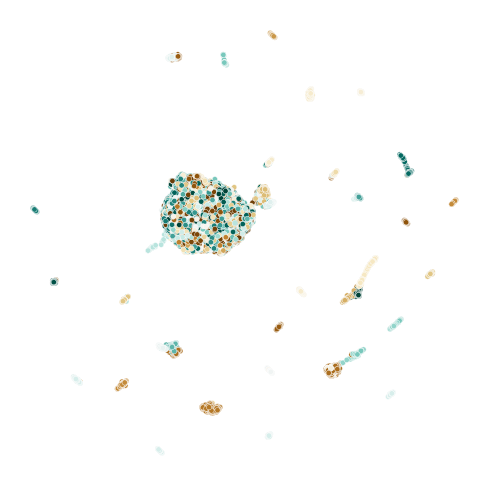}\\
\end{tabular}
\end{center}
\vspace{-4mm}
\caption{
    \emph{Feature embeddings.} UMAP visualization of the feature space, showing the impact of distribution shift and the effects of feature extractor alignment. The first column shows the features in S3, column two and three shows S5.  
    }
\label{fig:umap}
\vspace{-4mm}
\end{figure}
\arrayrulecolor{black}

\noindent\textbf{Effect of Data Granularity}
In our main experiments, we investigated the effectiveness of adapting the feature extractor to the full test source dataset, resulting in significant performance benefits, as shown in Table \ref{tab:main_results} and Figure \ref{fig:fig3}. 
However, in real drug discovery scenarios, data is generated in separate experimental batches, introducing batch variability. 
In order to process the data as it arrives, an online approach is desirable. 
Moreover, we expect batch-level variability to be particularly pronounced, as it involves controlling confounding environmental variables such as incubation time, reagent concentration, and device usage. 
Therefore, applying ODA at the batch level may prove advantageous compared to applying it across multiple batches.

To assess the performance of ODA and the baselines at different granularities, we conducted transfers from S3$\rightarrow$S11, S5$\rightarrow$S11, and S8$\rightarrow$S11, while varying the level of granularity at which the models were applied: plate, batch, and source. 
S11 consists of seven distinct plates belonging to four batches, and we applied ODA, along with the baseline models, separately for each plate and batch, resulting in a total of 11 settings (4 batches and 7 plates). 
We used the same setup as described earlier, with the mini-batch size reduced to 64 to accommodate the smaller dataset size when training on individual batches and plates. 
The results are illustrated in Figure \ref{fig:abl}.

Overall, ODA proves beneficial at all three levels of granularity (source, batch, and plate). 
Interestingly, aligning per batch yields the best performance, while plate-level alignment is slightly better or on par with full source training. 
This observation supports the known variability between batches, affirming the advantage of aligning within a group of similar variability. 
Additionally, since the same number of iterations is performed regardless of whether ODA is applied at the source or batch level, there is no additional cost associated with applying it to smaller subsets. 
In fact, it can be considered preferable as it facilitates easier parallelization.

\addtolength{\tabcolsep}{+5pt}  
\begin{table}[t]
\caption{Generalization performance (Accuracy) from S3 to S7 and S10, 
between DINO and MAE.
}
\vspace{-4mm}
\small
\input{tables/MAE-table}
\label{tab:MAE_table}
\vspace{-6mm}
\end{table}
\addtolength{\tabcolsep}{-5pt}  

\noindent\textbf{ODA using only a single plate or batch}
While aligning the feature space at the batch level proved to be the most effective non-cross-batch strategy, aligning features for each new batch or plate can be time-consuming (although it is still relatively low in time and cost compared to the experimental and imaging pipeline). 
To evaluate the feasibility of aligning the feature extractor when working with subsets of the data, we repeated the experiment described in the previous paragraphs. 
However, this time we evaluated each of the models on the full source dataset, rather than only on the subset it was aligned on. 
That is, we performed ODA on a single plate or batch, and applied this model to the rest of the out-of-domain data.
The results of this experiment are shown in Figure \ref{fig:abl}.
We observed a slight but noticeable drop in performance when models were evaluated on the full source dataset compared to when they were evaluated only on the subset they were aligned on. 
Nevertheless, the overall benefits of ODA were still maintained, as even aligning with a random plate led to substantial performance improvements over the supervised baseline.

\noindent\textbf{How ODA/CODA changes the feature space}
To gain a deeper understanding of how online alignment using ODA or CODA affects the feature space, we conducted a thorough analysis. 
We visualize the feature space of the dual model in Figure \ref{fig:umap} and performe Centered Kernel Alignment (CKA) \cite{cortes2014algorithms} analysis pre- and post-alignment in Figure \ref{fig:cka}.

In Figure \ref{fig:umap} we provide a UMAP of the embeddings of the CLS token to visualize the feature space. When transitioning from the in-domain to the out-of-domain setting (left and middle column), we observe a significant shift in the feature space for both the Dual-Model-Dino and Dual-Model-CB. 
In the in-domain setting (S3$\rightarrow$S3, left panels), clear structures are present.
Note that Dual-Model-DINO (S3$\rightarrow$S3) is over-clustered because it has picked up on batch effects, an undesirable property.
When applied to out-of-domain data (without adaptation, in the middle panels) S3$\rightarrow$S5
the models struggles to distinguish any mechanisms-of-action.
After applying ODA or CODA (right panels S3$\rightarrow$S5), the structure in the feature space is restored, allowing for better differentiation of MoA classes.

We further examined feature similarity within the feature extractor across different layers using CKA analysis. 
The results (Figure \ref{fig:cka}) revealed noticeable differences in feature representation pre- and post-alignment. 
This indicates that weights throughout the feature extractor are updated during alignment, particularly for high-level features. 
It is possible that biological variations between source setups are challenging to adapt to using low-level features.
Given the strong predictive performance of ODA and CODA models on out-of-domain data, as well as the good fit of the adapted features to the pre-trained task model, we focused on comparing the feature similarity between CODA before and after adaptation (S3$\rightarrow$S5).
We compare how closely aligned the pre- and post-aligned features resemble those learned in the in-domain setting.
Figure \ref{fig:cka} (top right) demonstrates a clear diagonal correlation between the post-alignment model and its S5 Dual-model-CB counterpart, suggesting that CODA training successfully learns features similar to those learned in the in-domain setting.
When examining the CLS feature similarity (Figure \ref{fig:cka} bottom right), the trend is clearer, and the strength of the similarity seems to be somewhat maintained, even at the high-level features.
Interestingly, without aligning the feature extractor, the CLS representations remained similar across different depths of the feature extractor (Figure \ref{fig:cka} bottom center), indicating that no high-level features are distinguishable in the new domain. This suggests that the domain shift is so severe that the learn features are no longer useful, potentially explaining why performance degradation is seen without alignment.

%% file: tables/main_table_TTT.tex
\begin{center}
\begin{tabular}{lcccccc@{\hskip 5pt}ccc@{\hskip 5pt}l}
\toprule
& &
\multicolumn{4}{c}{\textbf{Target}} & &
\\[0.5em] 
\textbf{Source} & &
S3 & 
S5 & 
S8 &
S11 & &
S7 &
S10 &
& Model type  (Set trained)
\\[0.5em] 
\cmidrule{2-6}\cmidrule{8-9}\cmidrule{11-11}
\multirow{6}{*}{S3} 
& \multicolumn{1}{c}{} &
\textit{ 39.0 $\pm$ 0.4} &
13.5 $\pm$ 1.3 &
15.6 $\pm$ 0.9 &
11.4 $\pm$ 1.0 & &
19.5 $\pm$ 0.8 &
12.0 $\pm$ 0.6 & &
Supervised
\\
& \multicolumn{1}{c}{} &
\textit{ 35.6 $\pm$ 0.2} &
9.2 $\pm$ 1.2 &
9.9 $\pm$ 0.1 &
7.1 $\pm$ 0.5 & &
14.8 $\pm$ 0.5 &
9.5 $\pm$ 0.8 & &
Dual-model-DINO
\\
& \multicolumn{1}{c}{} &
\textit{ 40.5 $\pm$ 0.4} &
14.3 $\pm$ 0.2 &
12.8 $\pm$ 1.1 &
9.0 $\pm$ 0.9 & &
17.9 $\pm$ 0.5 &
10.8 $\pm$ 0.7 & &
Dual-model-CB
\\
& \multicolumn{1}{c}{} &
- &
8.2 $\pm$ 1.4 &
9.0 $\pm$ 1.0 &
9.0 $\pm$ 2.8 & &
10.8 $\pm$ 3.3 &
7.6 $\pm$ 1.6 & &
TTT
\\
& \multicolumn{1}{c}{} &
- &
24.4 $\pm$ 0.6 &
25.4 $\pm$ 0.4 &
20.5 $\pm$ 1.3 & &
27.2 $\pm$ 0.6 &
17.3 $\pm$ 1.0 & &
ODA
\\
& \multicolumn{1}{c}{} &
- &
33.3 $\pm$ 0.3 &
33.9 $\pm$ 0.2 &
30.4 $\pm$ 0.6 & &
36.7 $\pm$ 0.3 &
26.7 $\pm$ 0.1 & &
CODA
\\[0.5em] 
\multirow{6}{*}{S5} 
& \multicolumn{1}{c}{} &
7.8 $\pm$ 0.2 &
\textit{ 35.9 $\pm$ 0.3} &
13.5 $\pm$ 0.5 &
5.9 $\pm$ 0.4 & &
11.7 $\pm$ 1.0 &
12.7 $\pm$ 0.2 & &
Supervised
\\
& \multicolumn{1}{c}{} &
5.2 $\pm$ 0.6 &
\textit{ 36.2 $\pm$ 0.6} &
13.6 $\pm$ 0.5 &
5.2 $\pm$ 0.7 & &
9.4 $\pm$ 0.4 &
11.9 $\pm$ 0.6 & &
Dual-model-DINO
\\
& \multicolumn{1}{c}{} &
6.0 $\pm$ 1.0 &
\textit{ 36.6 $\pm$ 0.7} &
14.9 $\pm$ 0.9 &
5.3 $\pm$ 1.0 & &
9.1 $\pm$ 0.5 &
12.6 $\pm$ 0.6 & &
Dual-model-CB
\\
& \multicolumn{1}{c}{} &
6.6 $\pm$ 1.6 &
- &
10.4 $\pm$ 1.8 &
5.9 $\pm$ 1.0 & &
9.2 $\pm$ 2.2 &
9.0 $\pm$ 1.2 & &
TTT
\\
& \multicolumn{1}{c}{} &
13.9 $\pm$ 1.1 &
- &
25.5 $\pm$ 0.4 &
14.9 $\pm$ 1.0 & &
23.4 $\pm$ 0.7 &
20.4 $\pm$ 0.7 & &
ODA
\\
& \multicolumn{1}{c}{} &
26.7 $\pm$ 1.3 &
- &
28.5 $\pm$ 1.1 &
21.2 $\pm$ 0.7 & &
27.6 $\pm$ 0.6 &
28.1 $\pm$ 0.5 & &
CODA
\\[0.5em] 
\multirow{6}{*}{S8} 
& \multicolumn{1}{c}{} &
9.5 $\pm$ 0.6 &
11.8 $\pm$ 1.4 &
\textit{ 31.0 $\pm$ 0.5} &
10.3 $\pm$ 1.5 & &
18.3 $\pm$ 0.4 &
12.0 $\pm$ 0.7 & &
Supervised
\\
& \multicolumn{1}{c}{} &
7.0 $\pm$ 0.7 &
12.0 $\pm$ 0.8 &
\textit{ 31.2 $\pm$ 0.9} &
8.4 $\pm$ 0.4 & &
13.0 $\pm$ 0.4 &
9.7 $\pm$ 0.3 & &
Dual-model-DINO
\\
& \multicolumn{1}{c}{} &
8.6 $\pm$ 0.5 &
12.0 $\pm$ 0.4 &
\textit{ 31.5 $\pm$ 1.9} &
9.3 $\pm$ 0.5 & &
13.3 $\pm$ 0.4 &
9.7 $\pm$ 0.6 & &
Dual-model-CB
\\
& \multicolumn{1}{c}{} &
5.1 $\pm$ 0.6 &
4.6 $\pm$ 0.5 &
- &
5.2 $\pm$ 0.5 & &
3.3 $\pm$ 0.2 &
4.7 $\pm$ 0.6 & &
TTT
\\
& \multicolumn{1}{c}{} &
14.1 $\pm$ 0.7 &
19.0 $\pm$ 0.6 &
- &
18.2 $\pm$ 0.3 & &
21.8 $\pm$ 0.8 &
17.2 $\pm$ 0.7 & &
ODA
\\
& \multicolumn{1}{c}{} &
20.1 $\pm$ 0.4 &
16.5 $\pm$ 0.9 &
- &
20.4 $\pm$ 1.0 & &
23.9 $\pm$ 1.1 &
19.9 $\pm$ 0.8 & &
CODA
\\[0.5em] 
\multirow{6}{*}{S11} 
& \multicolumn{1}{c}{} &
6.4 $\pm$ 1.0 &
6.6 $\pm$ 1.4 &
9.4 $\pm$ 0.7 &
\textit{ 32.4 $\pm$ 0.6} & &
7.9 $\pm$ 1.2 &
6.9 $\pm$ 1.3 & &
Supervised
\\
& \multicolumn{1}{c}{} &
4.7 $\pm$ 1.2 &
5.2 $\pm$ 1.4 &
6.6 $\pm$ 0.6 &
\textit{ 30.1 $\pm$ 0.3} & &
4.1 $\pm$ 0.5 &
4.4 $\pm$ 0.7 & &
Dual-model-DINO
\\
& \multicolumn{1}{c}{} &
4.5 $\pm$ 0.4 &
4.7 $\pm$ 0.7 &
6.9 $\pm$ 1.6 &
\textit{ 31.6 $\pm$ 0.8} & &
4.4 $\pm$ 1.0 &
4.4 $\pm$ 1.0 & &
Dual-model-CB
\\
& \multicolumn{1}{c}{} &
4.7 $\pm$ 0.2 &
4.9 $\pm$ 0.3 &
7.8 $\pm$ 1.4 &
- & &
7.0 $\pm$ 0.0 &
5.6 $\pm$ 0.6 & &
TTT
\\
& \multicolumn{1}{c}{} &
10.6 $\pm$ 0.2 &
14.8 $\pm$ 0.9 &
17.5 $\pm$ 1.1 &
- & &
17.7 $\pm$ 1.3 &
11.0 $\pm$ 0.3 & &
ODA
\\
& \multicolumn{1}{c}{} &
21.5 $\pm$ 0.5 &
16.5 $\pm$ 1.4 &
22.1 $\pm$ 0.9 &
- & &
23.8 $\pm$ 1.6 &
11.9 $\pm$ 1.2 & &
CODA
\\[0.5em] 
\bottomrule
\end{tabular}
\end{center}

%% file: tables/MAE-table.tex
\begin{center}
\begin{tabular}{lcc}
\toprule
Model type &
S3 $\rightarrow$ S7 &
S3 $\rightarrow$ S10 
\\
\midrule
Dual-model-MAE &
14.2 $\pm$ 0.8 &
8.9 $\pm$ 0.5  
\\
Dual-model-DINO &
\textbf{14.8 $\pm$ 0.5} &
\textbf{9.5 $\pm$ 0.8 }
\vspace{1mm}
\\
ODA-MAE &
17.8 $\pm$ 1.1 &
11.1 $\pm$ 0.4 
\\
ODA &
\textbf{27.2 $\pm$ 0.6} &
\textbf{17.3 $\pm$ 1.0}
\\

\bottomrule
\end{tabular}
\end{center}

%% file: sections/5_Discussion.tex
\section{Discussion}

Our empirical findings consistently support our initial expectations and highlight the limitations of standalone models in generalizing to new High Content Imaging (HCI) sources. 
In contrast, the dual model, with its adaptive capabilities, demonstrates a significant ability to reduce generalization gaps.
The dual model's bifurcated structure, consisting of separate feature extraction and task-specific components, facilitates a more straightforward adaptation to domain shifts. 
This design allows the task-specific features to remain intact while effectively adapting the feature extractor to new data characteristics.

The incorporation of CDCL into the self-supervised feature extraction process further strengthens the model's ability to mitigate batch effects, enhancing its overall robustness. 
CDCL ensures consistent representations across different batches, leading to substantial improvements in the model's generalization capabilities.

Our work highlights the critical importance of the training methodology employed in instructing the feature extractor via self-supervision. 
With the increasing development of innovative self-supervised methods, we anticipate the emergence of more advanced domain adaptation strategies in the near future. 
These strategies are expected to effectively address the current generalization gap in HCI, enabling more efficient and robust applications in this field.

Aside from the performance benefits demonstrated in our work, it is worth noting that methods like CODA that can adapt online to domain shifts are of critical importance in HCI and drug discovery, where sources of variation are high and unpredictable.
As such, online domain adaptation methods like CODA will be essential in mitigating these sources of variation that are ultimately beyond our control. 

%% file: sections/6_Conclusion.tex
\section{Conclusion}

\begin{figure}[t]
\begin{center}
\begin{tabular}{@{}c@{}c@{}c@{}}
\includegraphics[width=0.33\columnwidth]{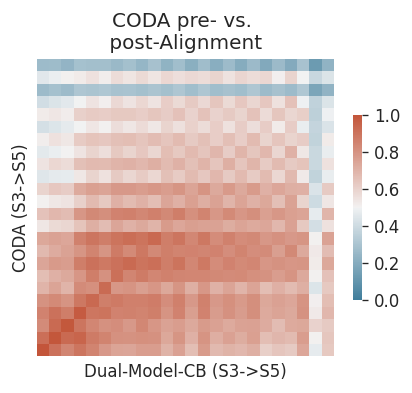}&
\includegraphics[width=0.33\columnwidth]{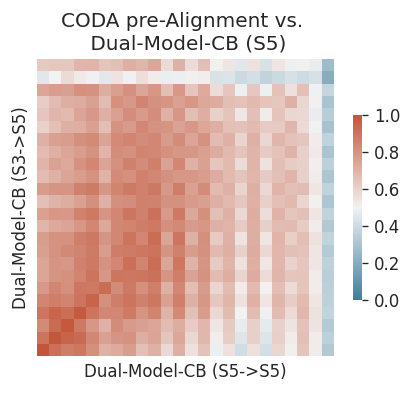}&
\includegraphics[width=0.33\columnwidth]{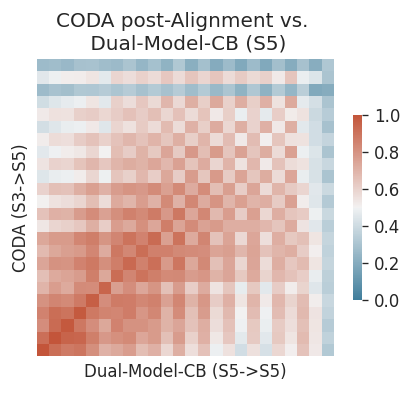}\\
\midrule
\includegraphics[width=0.33\columnwidth]{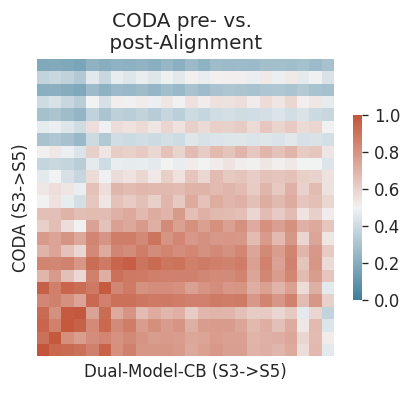}&
\includegraphics[width=0.33\columnwidth]{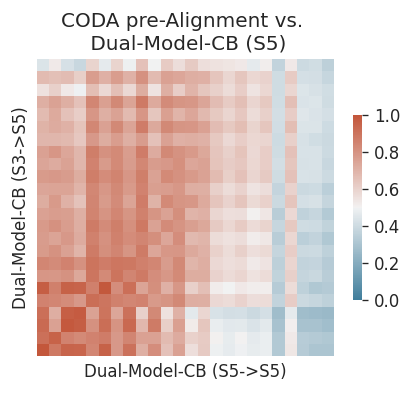}&
\includegraphics[width=0.33\columnwidth]{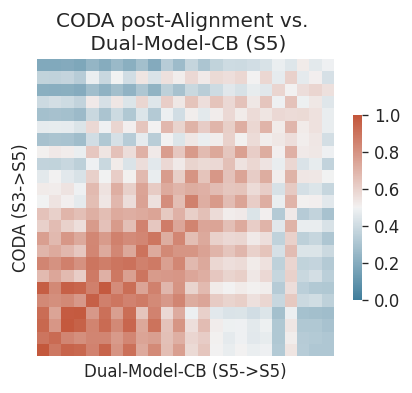}\\
\end{tabular}
\end{center}
\vspace{-4mm}
\caption{
    \emph{CKA.} Feature similarities in S5 for all tokens 
    (top) and CLS tokens (bottom)  of the feature extractor.
    (left) CODA before and after adaptation. (middle) CODA before adaptation vs. a Dual-Model-CB trained in S5. (right) CODA after adaptation vs. a Dual-Model-CB trained in S5.
    }
\label{fig:cka}
\vspace{-4mm}
\end{figure}

Our findings emphasize the limitations of standalone models when applied to novel High Content Imaging (HCI) data sources and highlight the effectiveness of the dual model approach in reducing generalization gaps. The dual model's bifurcated structure, comprising self-supervised feature extraction and task-specific components, enhances adaptability to new domain shifts. Furthermore, the integration of Cross-Domain Consistent Learning enhances the model's robustness and consistency across different batches, thereby improving its generalization capabilities. The advancement of sophisticated self-supervised methods is expected to drive progress in online domain adaptation strategies, ultimately addressing the prevailing generalization gap in high content imaging. Overall, our method offers a viable strategy to mitigate batch effects and distribution shifts caused by differences in experimental settings and apparatus, leading to improved generalization performance in the HCI domain.

%% file: sections/7_acknowledgements.tex
\paragraph{Acknowledgements.} This work was supported by the Wallenberg AI, Autonomous Systems and Software Program (WASP). We acknowledge the use of Berzelius computational resources provided by the Knut and Alice Wallenberg Foundation at the National Supercomputer Centre.
We would also like to thank the AWS Open Data Sponsorship Program for sponsoring the JUMP-CP data storage.

%% file: sections/9_Appendix.tex
\pagenumbering{Roman}
\setcounter{page}{1}
\twocolumn[
{\center\baselineskip 18pt
    \vskip .25in{\Large\bf
    Supplementary Material for Bridging Generalization Gaps in High Content Imaging Through Online Self-Supervised Domain Adaptation \par
}\vskip .52in}
]
\begin{appendices}

\begin{figure}[t]
\begin{center}
\begin{tabular}{@{}c@{}}
\includegraphics[width=1\columnwidth]{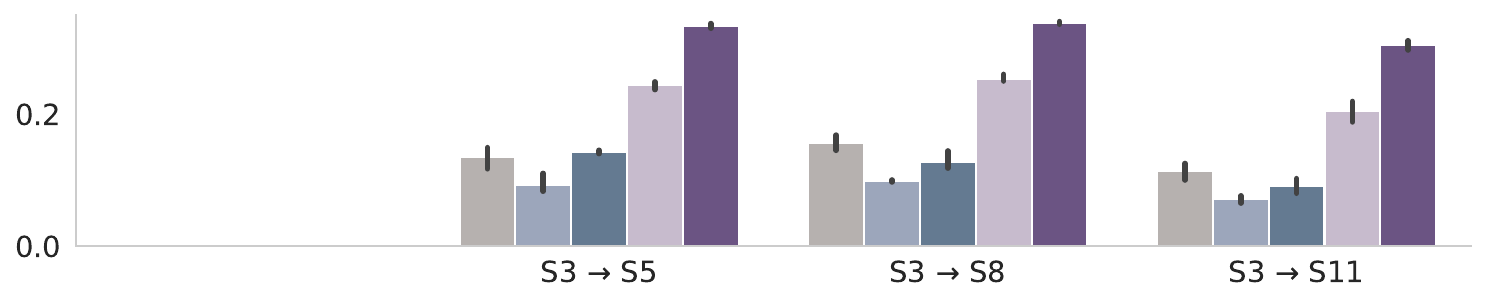}
\vspace{-3mm}\\
\includegraphics[width=1\columnwidth]{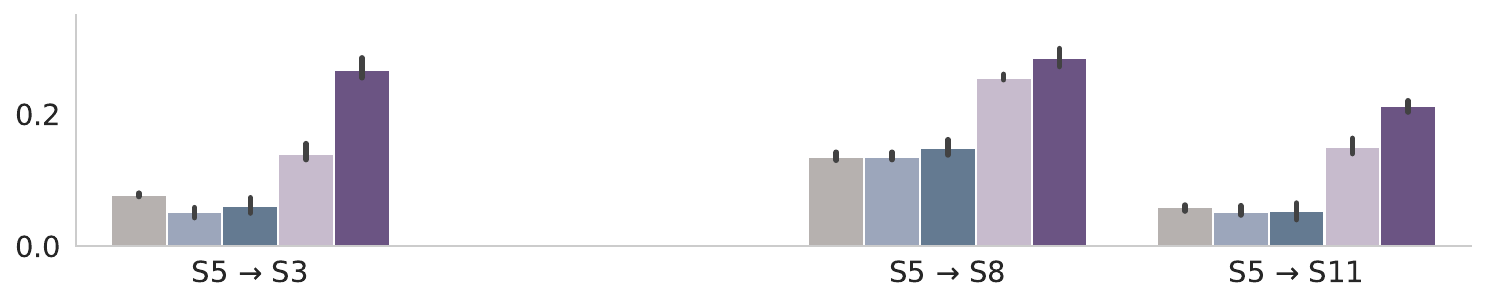}
\vspace{-3mm}\\
\includegraphics[width=1\columnwidth]{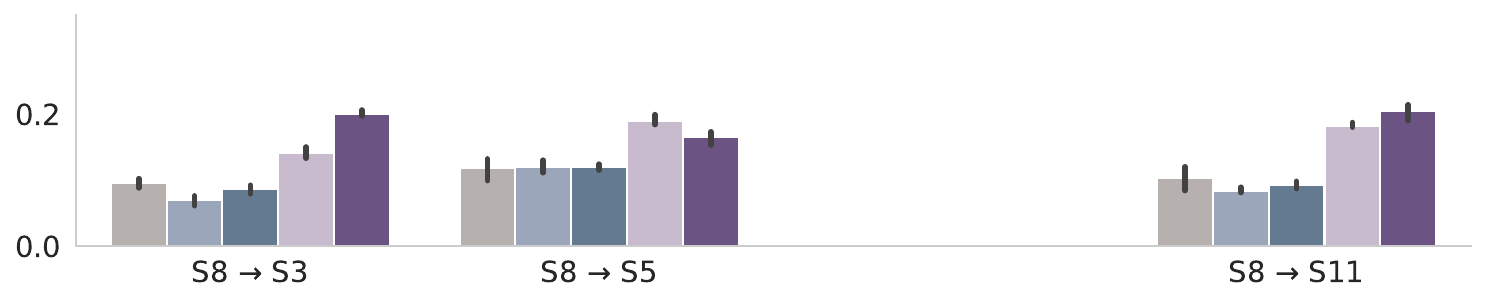}
\vspace{-3mm}\\
\includegraphics[width=1\columnwidth]{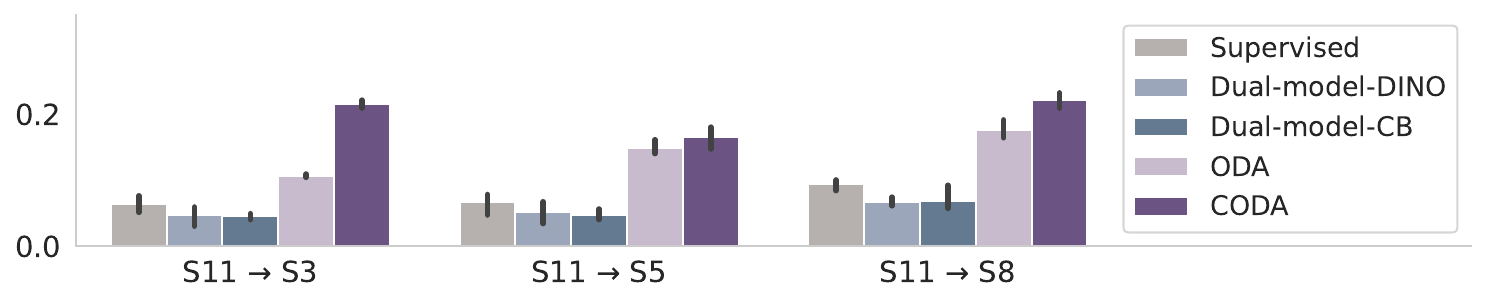}\\
\end{tabular}
\end{center}
\vspace{-4mm}
\caption{
    \emph{Out-of-domain test performance (Acc.).} We compare the various learning setups trained on data from one source and applied to some target data, without access to labels from the target.}
\label{fig:fig3}
\vspace{-4mm}
\end{figure}

\section{Appendix overview}
Here, we provide further details and results from the experiments carried out in this work. 
In Section \ref{appx:aux_res} we present supplementary results and figures derived from the experiments discussed in the main text and in Section \ref{app:more_data} we provide additional information regarding the datasets and sources used in this study.

\subsection{Additional results}
\label{appx:aux_res}
Here we provide auxiliary information, results and figures from the experiments run and data used in this work.

In Table \ref{tab:main_results_f1}, re report results from the same experiments, discussed in the main text and reported in Table \ref{tab:main_results}.
We report F1 scores, corroborating the results reported in the main text when using Accuracy.
The main results reported for in Table \ref{tab:main_results} (excluding TTT), are also reported in bar-chart format in Figure \ref{fig:fig1} -- clearly visualizing the performance boosts of ODA and CODA compared to the baseline.
In Table \ref{tab:TTT-draft} additional experiments containing the performance of the DUAL-Model-MAE and TTT are reported between each of the sources used.

\addtolength{\tabcolsep}{-5pt}  
\begin{table}[t]
\caption{Generalization performance across target sources (\textit{F1}.)
}
\vspace{-4mm}
\tiny
\input{tables/main_table_f1}
\label{tab:main_results_f1}
\vspace{-4mm}
\end{table}
\addtolength{\tabcolsep}{+5pt}  

\addtolength{\tabcolsep}{-5pt}  
\begin{table}[t]
\caption{Comparison of MAE based dual-model and Test-Time Training performance (accuracy).
}
\vspace{-4mm}
\tiny
\input{tables/TTT_table}
\label{tab:TTT-draft}
\vspace{-4mm}
\end{table}
\addtolength{\tabcolsep}{+5pt}

\subsection{Detailed Data Description} 
\label{app:more_data}

As described in section \ref{sec:experimental_setup} in the main text, the primary experiments focus on a subset of the data from four (anonymized) partners within the JUMP-CP consortium \cite{JUMP_CP_DATASET}. The data of those sources along with two additional test sources are described in Table \ref{tab:sample_setting}. Here we include further information about these sources. Starting with the four primary sources that were selected, based on containing the largest subsets of data from each of the different microscope types used, thus, providing the most diverse set of data sources:

\begin{itemize}
\vspace{-1.5mm}
    \item \textbf{S3} contains 25 plates, totaling 9,600 unique wells and 85,409 images in total, belonging to 13 distinct experimental batches. These were captured using the Opera Phoenix microscope in widefield mode, using laser excitation and a 20X/1 NA objective.
    \vspace{-1.5mm}
    \item \textbf{S5}  contains 24 plates, totaling 9,216 unique wells and 82,256 images in total, belonging to 23 distinct experimental batches. These were captured using the CV8000 confocal microscope, using laser excitation and a 20X/0.75 NA objective.
    \vspace{-1.5mm}
    \item \textbf{S8} contains 4 plates, totaling 1,536 unique wells and 13,824 images in total, belonging to 4 distinct experimental batches. These were captured using the ImageExpress Micro confocal microscope, using LED excitation and a 20X/0.75 NA objective.
    \vspace{-1.5mm}
    \item \textbf{S11} contains 7 plates, totaling 2,688 unique wells and 23,373 images in total, belonging to 4 distinct experimental batches. These were captured using the Operetta widefield microscope, using LED excitation and a 20X/1 NA objective.
\end{itemize}

Two additional sources were also used for auxiliary testing. Both use similar microscope setups to that used by S5, allowing comparison of generalization performance between models trained and tested in sources with similar imaging setups.

\begin{itemize}
\vspace{-1.5mm}
    \item \textbf{S7} contains 7 plates, totaling 2,688 unique wells and 24,192 images in total, belonging to 7 distinct experimental batches. These were captured using the CV7000 confocal microscope, using laser excitation and a 20X/0.75 NA objective. 
    \vspace{-1.5mm}
    \item \textbf{S10} contains 6 plates, totaling 2,304 unique wells and 13,812 images in total, belonging to 6 distinct experimental batches. These were captured using the CV8000 confocal microscope, using laser excitation and a 20X/0.75 NA objective.
\end{itemize}

\end{appendices}

%% file: tables/main_table_f1.tex
\begin{center}
\begin{tabular}{lcccccc@{\hskip 5pt}ccc@{\hskip 5pt}l}
\toprule
& &
\multicolumn{4}{c}{\textbf{Target}} & &
\\[0.5em] 
\textbf{Source} & &
S3 & 
S5 & 
S8 &
S11 & &
S7 &
S10 &
& Model type  (Set trained)
\\[0.5em] 
\cmidrule{2-6}\cmidrule{8-9}\cmidrule{11-11}
\multirow{5}{*}{S3} 
& \multicolumn{1}{c}{} &
- &
10.9 $\pm$ 1.9 &
11.0 $\pm$ 0.5 &
8.1 $\pm$ 0.3 &
&
15.3 $\pm$ 0.5 &
9.0 $\pm$ 1.1 & &
Supervised
\\
& \multicolumn{1}{c}{} &
- &
6.2 $\pm$ 0.7 &
6.5 $\pm$ 0.5 &
4.9 $\pm$ 0.5 &
&
10.2 $\pm$ 0.4 &
6.4 $\pm$ 0.2 & &
Dual-model-DINO
\\
& \multicolumn{1}{c}{} &
- &
10.6 $\pm$ 0.5 &
8.3 $\pm$ 0.8 &
6.0 $\pm$ 0.5 &
&
13.2 $\pm$ 0.7 &
7.9 $\pm$ 0.8 & &
Dual-model-CB
\\
& \multicolumn{1}{c}{} &
- &
6.4 $\pm$ 1.3 &
5.3 $\pm$ 1.9 &
5.6 $\pm$ 2.7 & &
7.6 $\pm$ 3.5 &
5.2 $\pm$ 2.2 & &
TTT
\\
& \multicolumn{1}{c}{} &
- &
22.2 $\pm$ 0.7 &
23.3 $\pm$ 0.1 &
19.1 $\pm$ 1.7 &
&
23.5 $\pm$ 0.6 &
15.6 $\pm$ 0.7 & &
ODA
\\
& \multicolumn{1}{c}{} &
- &
30.7 $\pm$ 0.6 &
31.9 $\pm$ 0.5 &
28.6 $\pm$ 1.3 &
&
32.8 $\pm$ 0.4 &
24.4 $\pm$ 0.3 & &
CODA
\\[0.5em] 
\multirow{5}{*}{S5} 
& \multicolumn{1}{c}{} &
5.5 $\pm$ 0.8 &
- &
9.4 $\pm$ 0.4 &
2.5 $\pm$ 0.3 &
&
7.1 $\pm$ 0.9 &
9.4 $\pm$ 0.0 & &
Supervised
\\
& \multicolumn{1}{c}{} &
3.6 $\pm$ 0.3 &
- &
10.9 $\pm$ 0.5 &
2.6 $\pm$ 0.2 &
&
5.6 $\pm$ 0.3 &
8.9 $\pm$ 0.7 & &
Dual-model-DINO
\\
& \multicolumn{1}{c}{} &
3.5 $\pm$ 0.6 &
- &
11.0 $\pm$ 0.8 &
2.9 $\pm$ 0.1 &
&
5.2 $\pm$ 0.1 &
9.9 $\pm$ 0.6 & &
Dual-model-CB
\\
& \multicolumn{1}{c}{} &
4.1 $\pm$ 1.2 &
- &
7.2 $\pm$ 1.6 &
3.6 $\pm$ 0.5 & &
6.0 $\pm$ 1.7 &
6.6 $\pm$ 1.4 & &
TTT
\\
& \multicolumn{1}{c}{} &
13.4 $\pm$ 0.7 &
- &
22.0 $\pm$ 0.6 &
13.1 $\pm$ 0.5 &
&
19.5 $\pm$ 0.9 &
18.0 $\pm$ 0.5 & &
ODA
\\
& \multicolumn{1}{c}{} &
23.3 $\pm$ 1.0 &
- &
25.0 $\pm$ 1.1 &
18.0 $\pm$ 0.9 &
&
23.1 $\pm$ 0.4 &
24.0 $\pm$ 0.5 & &
CODA
\\[0.5em] 
\multirow{5}{*}{S8} 
& \multicolumn{1}{c}{} &
5.6 $\pm$ 0.2 &
9.1 $\pm$ 0.6 &
- &
6.2 $\pm$ 1.3 &
&
14.1 $\pm$ 0.3 &
9.6 $\pm$ 0.3 & &
Supervised
\\
& \multicolumn{1}{c}{} &
3.4 $\pm$ 0.2 &
8.8 $\pm$ 0.3 &
- &
4.7 $\pm$ 0.3 &
&
9.2 $\pm$ 0.3 &
8.1 $\pm$ 0.1 & &
Dual-model-DINO
\\
& \multicolumn{1}{c}{} &
4.8 $\pm$ 0.4 &
8.9 $\pm$ 0.3 &
- &
5.8 $\pm$ 0.2 &
&
9.2 $\pm$ 0.2 &
8.5 $\pm$ 0.5 & &
Dual-model-CB
\\
& \multicolumn{1}{c}{} &
2.4 $\pm$ 0.1 &
1.4 $\pm$ 0.1 &
- &
3.0 $\pm$ 0.1 & &
1.5 $\pm$ 0.0 &
2.3 $\pm$ 0.1 & &
TTT
\\
& \multicolumn{1}{c}{} &
10.1 $\pm$ 1.0 &
13.4 $\pm$ 0.3 &
- &
15.0 $\pm$ 0.6 &
&
16.2 $\pm$ 1.5 &
13.5 $\pm$ 0.6 & &
ODA
\\
& \multicolumn{1}{c}{} &
16.1 $\pm$ 0.5 &
13.9 $\pm$ 1.3 &
- &
17.3 $\pm$ 1.0 &
&
19.3 $\pm$ 1.1 &
16.7 $\pm$ 0.5 & &
CODA
\\[0.5em] 
\multirow{5}{*}{S11} 
& \multicolumn{1}{c}{} &
4.1 $\pm$ 0.5 &
3.7 $\pm$ 0.2 &
6.3 $\pm$ 0.4 &
- &
&
5.0 $\pm$ 0.3 &
4.2 $\pm$ 0.3 & &
Supervised
\\
& \multicolumn{1}{c}{} &
1.8 $\pm$ 0.3 &
1.6 $\pm$ 0.5 &
3.5 $\pm$ 0.1 &
- &
&
2.3 $\pm$ 0.3 &
1.8 $\pm$ 0.3 & &
Dual-model-DINO
\\
& \multicolumn{1}{c}{} &
2.1 $\pm$ 0.4 &
2.1 $\pm$ 0.3 &
4.3 $\pm$ 0.5 &
- &
&
2.8 $\pm$ 0.6 &
2.4 $\pm$ 0.4 & &
Dual-model-CB
\\
& \multicolumn{1}{c}{} &
2.4 $\pm$ 0.6 &
2.4 $\pm$ 0.8 &
4.4 $\pm$ 1.5 &
- & &
3.8 $\pm$ 0.8 &
2.9 $\pm$ 0.8 & &
TTT
\\
& \multicolumn{1}{c}{} &
7.9 $\pm$ 0.5 &
10.4 $\pm$ 0.7 &
13.8 $\pm$ 1.2 &
- &
&
13.2 $\pm$ 0.7 &
6.8 $\pm$ 0.1 & &
ODA
\\
& \multicolumn{1}{c}{} &
18.2 $\pm$ 0.9 &
13.2 $\pm$ 0.9 &
19.4 $\pm$ 0.7 &
- &
&
19.7 $\pm$ 1.1 &
8.7 $\pm$ 0.6 & &
CODA
\\[0.5em] 
\bottomrule
\end{tabular}
\end{center}

%% file: tables/TTT_table.tex
\begin{center}
\begin{tabular}{lcccccc@{\hskip 5pt}ccc@{\hskip 5pt}l}
\toprule
& &
\multicolumn{4}{c}{\textbf{Target}} & &
\\[0.5em] 
\textbf{Source} & &
S3 & 
S5 & 
S8 &
S11 & &
S7 &
S10 &
& Model type  (Set trained)
\\[0.5em] 
\cmidrule{2-6}\cmidrule{8-9}\cmidrule{11-11}
\multirow{2}{*}{S3} 
& \multicolumn{1}{c}{} &
- &
11.9 $\pm$ 3.5 &
12.8 $\pm$ 3.3 &
12.1 $\pm$ 1.0 & &
14.9 $\pm$ 2.6 &
9.4 $\pm$ 1.5 & &
Dual-model-MAE
\\
& \multicolumn{1}{c}{} &
- &
8.2 $\pm$ 1.4 &
9.0 $\pm$ 1.0 &
9.0 $\pm$ 2.8 & &
10.8 $\pm$ 3.3 &
7.6 $\pm$ 1.6 & &
TTT
\\[0.5em] 
\multirow{2}{*}{S5} 
& \multicolumn{1}{c}{} &
7.6 $\pm$ 2.2 &
- &
13.9 $\pm$ 0.4 &
6.8 $\pm$ 1.3 & &
10.2 $\pm$ 0.8 &
12.4 $\pm$ 1.3 & &
Dual-model-MAE
\\
& \multicolumn{1}{c}{} &
6.6 $\pm$ 1.6 &
- &
10.4 $\pm$ 1.8 &
5.9 $\pm$ 1.0 & &
9.2 $\pm$ 2.2 &
9.0 $\pm$ 1.2 & &
TTT
\\[0.5em] 
\multirow{2}{*}{S8} 
& \multicolumn{1}{c}{} &
9.1 $\pm$ 0.4 &
11.0 $\pm$ 0.1 &
- &
10.7 $\pm$ 0.1 & &
14.4 $\pm$ 0.4 &
10.5 $\pm$ 0.4 & &
DUAL-model-MAE
\\
& \multicolumn{1}{c}{} &
5.1 $\pm$ 0.6 &
4.6 $\pm$ 0.5 &
- &
5.2 $\pm$ 0.5 & &
3.3 $\pm$ 0.2 &
4.7 $\pm$ 0.6 & &
TTT
\\[0.5em] 
\multirow{2}{*}{S11} 
& \multicolumn{1}{c}{} &
7.9 $\pm$ 0.6 &
8.2 $\pm$ 0.1 &
10.6 $\pm$ 1.1 &
- & &
9.4 $\pm$ 0.2 &
7.7 $\pm$ 0.2 & &
DUAL-model-MAE
\\
& \multicolumn{1}{c}{} &
4.7 $\pm$ 0.2 &
4.9 $\pm$ 0.3 &
7.8 $\pm$ 1.4 &
- & &
7.0 $\pm$ 0.0 &
5.6 $\pm$ 0.6 & &
TTT
\\[0.5em] 
\bottomrule
\end{tabular}
\end{center}